# Applications of Federated Learning in Manufacturing: Identifying the Challenges and Exploring the Future Directions with Industry 4.0 and 5.0 Visions

**Farzana Islam, Ahmed Shoyeb Raihan, Imtiaz Ahmed**
Department of Industrial and Management Systems Engineering
West Virginia University
Morgantown, West Virginia, USA
fi00003@mix.wvu.edu, ar00065@mix.wvu.edu, imtiaz.ahmed@mail.wvu.edu

## Abstract

In manufacturing settings, data collection and analysis are often a time-consuming, challenging, and costly process. It also hinders the use of advanced machine learning and data-driven methods which require a substantial amount of offline training data to generate good results. It is particularly challenging for small manufacturers who do not share the resources of a large enterprise. Recently, with the introduction of the Internet of Things (IoT), data can be collected in an integrated manner across the factory in real-time, sent to the cloud for advanced analysis, and used to update the machine learning model sequentially. Nevertheless, small manufacturers face two obstacles in reaping the benefits of IoT: they may be unable to afford or generate enough data to operate a private cloud, and they may be hesitant to share their raw data with a public cloud. Federated learning (FL) is an emerging concept of collaborative learning that can help small-scale industries address these issues and learn from each other without sacrificing their privacy. It can bring together diverse and geographically dispersed manufacturers under the same analytics umbrella to create a win-win situation. However, the widespread adoption of FL across multiple manufacturing organizations remains a significant challenge. This study aims to review the challenges and future directions of applying federated learning in the manufacturing industry, with a specific emphasis on the perspectives of Industry 4.0 and 5.0.

## 1. Introduction

Federated Learning (FL) is a decentralized machine learning technique that enables multiple parties to collaboratively train machine learning (ML) models while keeping their data securely stored on their local devices. The concept of FL was first introduced in 2016 by Google researchers in a paper titled "Communication-Efficient Learning of Deep Networks from Decentralized Data" (McMahan et al., 2017). The motivation behind this work was to address the growing concern for privacy in machine learning, where data from individual users are often centralized and stored in large data centers. This was where FL came into play, offering a new way of training ML models in a decentralized manner, where data remains on the user's device, and only the model parameters are transmitted to the aggregator. With the rapid increase in the amount of time series data generated by devices, the manufacturing industry is shifting towards real-time/online analytics as it offers improved predictive maintenance, better quality control, enhanced process optimization, and thus overall increased efficiency. However, this online learning needs a very large dataset for the initial training which is often difficult for the small and medium-sized enterprise (SME) industry to obtain due to its small-scale production. This problem can easily be overcome by collaborating with other entities in the SME industry by sharing a decentralized database. However, this cannot protect the individual company data from privacy and security threats which is why FL is the perfect candidate for this situation. Although FL promises to address all these issues and the manufacturing industry is heading towards smart manufacturing where a noticeable amount of increase has been observed in installing IoT devices, implementation of FL is very rare in the manufacturing industry. This has inspired us to figure out the challenges behind the implementation of FL in the manufacturing industry. In Figure 1, key components of smart manufacturing have been shown to demonstrate the basic FL architecture composed of cloud and edge computing. The objective of this study is to explore the challenges associated with implementing

FL in the manufacturing industry from the perspectives of Industry 4.0 and Industry 5.0. Alongside this exploration, we propose solutions to overcome these challenges and maximize the benefits of FL in manufacturing.

The rest of the paper unfolds as follows. Section 2 provides a brief review of literature of the related works and discusses the research gaps of FL. In section 3, different types of FL frameworks are illustrated with their characteristics. The applications of FL in manufacturing industry in the context of Industry 4.0/5.0 has been discussed in section 4. Section 5 explores the current challenges related to the applications of FL in manufacturing and discusses potential solutions to tackle them. Finally, we conclude our work in section 6.

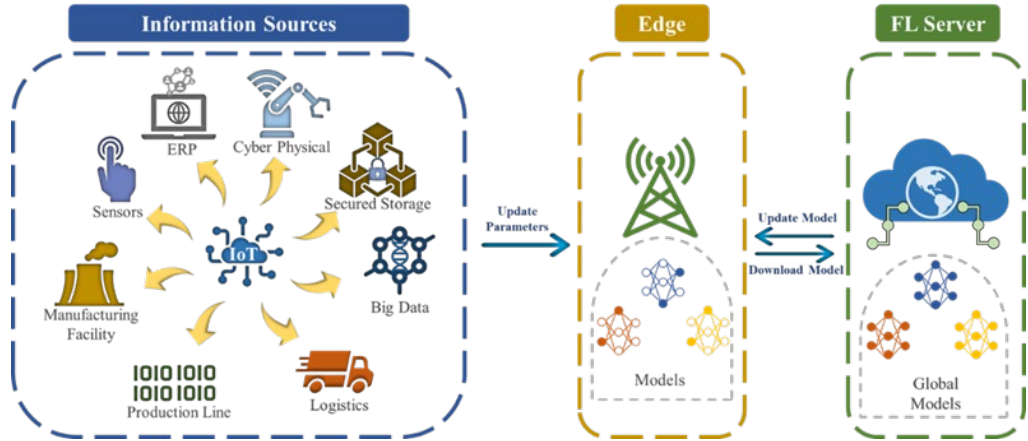

Figure 1: Framework of FL model for smart manufacturing system

## 2. Related Works

Federated Learning (FL) is a revolutionary method that allows multiple users to collaboratively train a global machine learning model based on local weights derived from their individual data without revealing sensitive information. Initially introduced by Google in 2016, this method is repeated to achieve high accuracy in the global model (McMahan et al., 2017). Unlike traditional centralized machine learning methods, FL reduces the systemic privacy risks and costs associated with data exchange by adhering to the principles of data privacy and protection stipulated in the General Data Protection Regulation (GDPR) (Hegiste et al., 2022). There has been a significant number of research works on FL since its inception. In many of these works, a comparative study of FL is conducted with distributed learning, parallel learning, and deep learning (Aledhari et al., 2020). Several studies mainly focus on the comparison of Deep Learning and FL due to the widespread use of Deep Neural Networks (DNNs) in the field of machine learning (Ben-Nun & Hoefler, 2019; L. Liu et al., 2020; Poirot et al., 2019; Vepakomma et al., 2018). DNNs have gained popularity in various applications and have shown promising results. However, DNNs also have limitations that pose challenges for their integration into FL. The study by Li et al. discusses the future directions and feasibility of Federated Learning (FL) in terms of data privacy and protection. They also conduct comparisons of different FL systems (Q. Li et al., 2023). In another study, the authors conducted a survey on techniques for FL, with a specific focus on personalization techniques (Kulkarni et al., 2020). In their work, the authors elaborate on the reasons why it is crucial to consider personalization in FL. The study by Geiping et al. (Geiping et al., 2020) attempts to address the question of the vulnerability of privacy in FL, which is crucial for the successful implementation of FL. Additionally, the article not only explores FL in general but also specifically focuses on the utilization of FL for image data recovery. While previous literature has explored the benefits of FL compared to classical machine learning in terms of privacy protection and convergence efficiency (Asad et al., 2021), fewer works have identified the potential barriers to its implementation in the manufacturing industry. This paper aims to address this gap by reviewing existing literature to classify the challenges of applying FL in the manufacturing industry and provide suggestions for future improvements (Kairouz et al., 2021; Zhang et al., 2021). Previous works have suggested the application of FL in manufacturing (Hegiste et al., 2022; Leng et al., 2021; Wong & Kim, 2017) but none have identified potential barriers that might interrupt the application of FL in the manufacturing industry.

## 3. Classification of FL

The classification of FL has its roots in the evolution of machine learning and the growing need for privacy and security of data. To address the limitations of traditional machine learning models in handling complex and large-scale data, FL was classified into three types to deal with large and complex data, while preserving privacy and reducing computational costs (Zhang et al., 2021). The following table discusses the three types of FL along with their definition and characteristics. Figure 2 illustrates the frameworks of Horizontal FL, Vertical FL, and Transfer FL, respectively.

Table 1: Classification of Federated Learning

| Type of Federated Learning | Definition | Characteristics |
|---|---|---|
| Horizontal Federated Learning | In horizontal FL, data samples belonging to the same data distribution are distributed across different devices | <ul><li>Data samples belong to the same distribution.</li><li>Data distribution is identical across devices.</li><li>Aimed at increasing accuracy by aggregating data from multiple devices.</li><li>Example: Credit card transactions from different banks</li></ul> |
| Vertical Federated Learning | In vertical FL, data samples belonging to different data distributions are present on the same device | <ul><li>Data samples belong to different distributions.</li><li>Data distribution varies across devices.</li><li>Aimed at preserving privacy by training models on encrypted local data.</li><li>Example: Healthcare data from different patients on the same device</li></ul> |
| Transfer Federated Learning | Transfer FL is used when the data distributions among the devices are different | <ul><li>Data samples belong to different distributions.</li><li>Data distributions vary across devices.</li><li>Aimed at transferring knowledge learned from one distribution to improve the learning performance on another distribution.</li><li>Example: Sentiment analysis model trained on English data and then transferred to Spanish data</li></ul> |

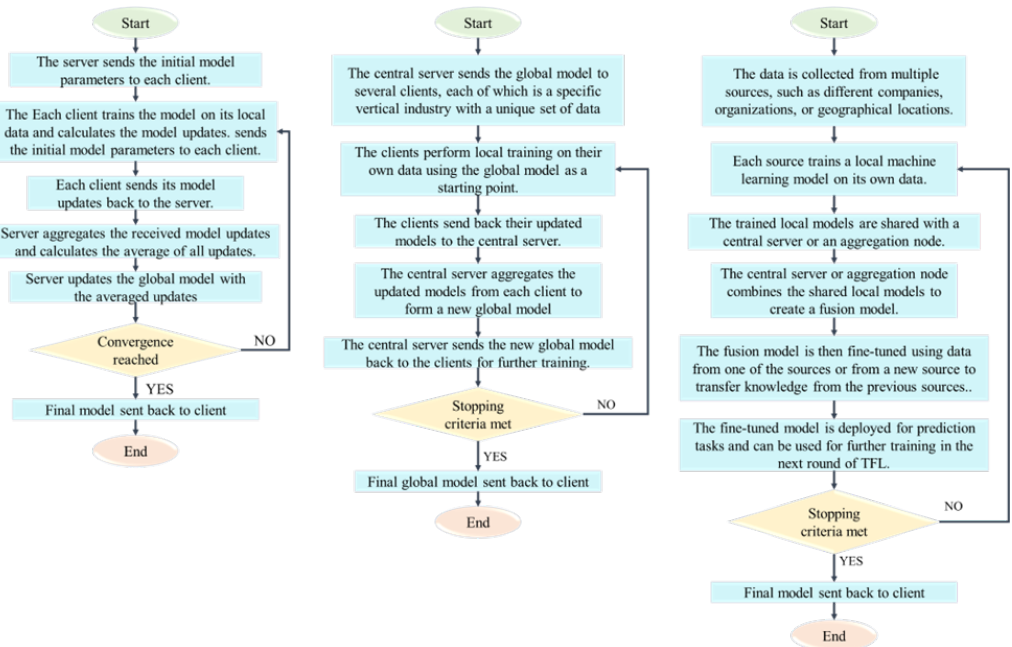

Figure 2: Frameworks of Horizontal FL (left), Vertical FL (middle) and Transfer FL (right)

## 4. FL in Manufacturing

FL in a manufacturing environment works by enabling multiple parties, such as suppliers, customers, or even machines on the shop floor, to collaboratively train a machine learning model using their local data. The local data stays on the individual devices and only the model updates are sent to a central server, reducing privacy concerns. The central server then aggregates the model updates and broadcasts the improved global model back to the individual devices, which repeat the process. In figure 2(a), the flow charts of the basic framework of FL have been illustrated.

### 4.1. Applications of FL in Manufacturing Industry

Throughout time, the manufacturing sector has experienced considerable transformations due to technological revolutions including big data, Industry 4.0, the Internet of Things, cloud computing, and artificial intelligence (Khosravi et al., 2023). FL can help the manufacturing industry by enabling collaboration among multiple parties to share data, create more accurate models, leverage expertise, and develop innovative solutions. This can enhance competitiveness, adaptability, and market position. Additionally, FL offers data privacy and security benefits as sensitive data is not transferred to a central server. This can be leveraged in the manufacturing industry for SMEs in both inter-company and intra-company collaboration scenarios. In intra-company collaboration, multiple departments or teams within the same SME can work together using FL to build and train models, improving the efficiency and accuracy of the manufacturing process. In inter-company collaboration, multiple SMEs can work together to train a global model using their local data while keeping it private, resulting in an accurate and efficient model. This will allow SMEs to leverage the benefits of collaborative machine learning without having to invest in expensive data storage and computational resources. Figure 3 shows the application of FL in the manufacturing industry where it has a huge potential for transforming the Quality Control sector along with the others illustrated in the diagram. At present, the entire manufacturing industry is significantly affected by the fourth industrial revolution. Consequently, it is beyond any doubt that when the fifth industrial revolution begins, the manufacturing industry will go through major changes once again. Therefore, in the following sections, we discuss the applications of FL from the perspective of Industry 4.0 and Industry 5.0.

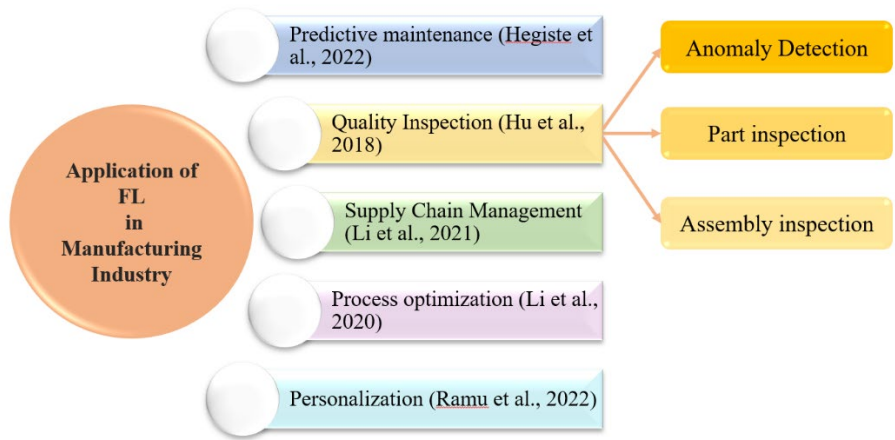

Figure 3: Application of FL in the manufacturing industry

### 4.2. Industry 4.0 and FL

Industry 4.0 refers to the fourth industrial revolution, which is characterized by the integration of advanced digital technologies into all aspects of the manufacturing process, including production, supply chain management, and customer engagement. Key enablers of Industry 4.0 include the internet of things (IoT), big data and analytics, artificial intelligence, additive manufacturing, cloud computing, and cybersecurity. As Industry 4.0 technologies become increasingly prevalent and substantial datasets become more accessible, scientists and researchers are increasingly dependent on data-driven methodologies that harness machine learning frameworks to effectively tackle diverse challenges (Raihan & Ahmed, 2019). Some notable applications of FL in Industry 4.0 are discussed in the following. In the context of Industry 4.0, where data-driven decision-making is crucial for optimizing production processes, FL can play a significant role. It enables collaborative model training across different factories, allowing them to collectively improve production models while ensuring data privacy. For instance, factories can share valuable insights and learnings from their local data, leading to enhanced predictive maintenance, process optimization, and quality control, without compromising the confidentiality of proprietary data. FL can facilitate the training of predictive maintenance models across multiple organizations that possess similar types of machinery, without the need to share sensitive maintenance data. This approach can yield more accurate and robust predictive maintenance models, resulting in reduced downtime, prolonged machinery lifespan, and improved productivity. In the vast field of supply chain, FL can enable collaborative training of machine learning models for supply chain optimization, including demand forecasting, inventory optimization, and logistics optimization, across various stakeholders in the supply chain while ensuring data privacy. This collaborative approach can lead to optimized supply chain operations, reduced inventory costs, and improved customer service levels, benefiting all participants in the supply chain ecosystem. In addition, FL can be applied in training machine learning models for energy management in diverse factories or energy-intensive processes, facilitating the exchange of insights and learnings while safeguarding sensitive energy consumption data. Such collaborative approach can yield more accurate energy consumption prediction models and optimized energy management strategies, resulting in reduced energy costs and enhanced sustainability practices. Furthermore, FL can enable collaborative training of machine learning models for product quality improvement across multiple organizations, allowing them to collectively identify defects, anomalies, or quality issues in products while safeguarding the confidentiality of their proprietary product data. This collaborative approach can lead to enhanced product quality, reduced product recalls, and improved customer satisfaction, benefiting all organizations involved in the collaborative learning process.

### 4.3. Industry 5.0 and FL

Industry 5.0 is an emerging concept that builds upon the industry 4.0 framework and places a greater emphasis on the role of human workers in manufacturing. Industry 5.0 aims to create a more collaborative and flexible manufacturing environment, where advanced technologies are integrated with human skills and expertise. It will adopt the key enablers of Industry 4.0 such as IoT, big data analytics, and artificial intelligence, and enable the industry to integrate human workers with machines. It aims to achieve high levels of collaboration, flexibility, and agility. Some of the applications of FL in Industry 5.0 are discussed in this section. In the context of Industry 5.0, which emphasizes collaboration between humans and robots, FL can enable collaborative training of machine learning models directly on robots or other edge devices, allowing them to learn and adapt locally without transmitting data to a central server. This approach can result in more personalized and context-aware robot behaviors, improved safety measures, and enhanced productivity in human-robot collaborative scenarios. Personalized manufacturing, a concept more applicable to Industry 5.0, is also greatly impacted by FL. FL can facilitate the customization of manufacturing processes and products based on individual preferences or requirements, enabling organizations to collaboratively train machine learning models using user data while preserving privacy. The insights gained from these models can then be used to customize products, processes, or services according to individual needs, resulting in more personalized and customer-centric manufacturing approaches. The use of technologies such as Augmented Reality (AR) and Virtual Reality (VR) will be more frequent and prevalent in the fifth industrial revolution. Organizations can collaborate to train models that improve object recognition, scene understanding, or gesture recognition in AR/VR environments, while ensuring user privacy is preserved. This approach can result in more immersive and realistic AR/VR experiences in industrial settings, enhancing the usability and effectiveness of these technologies in Industry 5.0 scenarios. Besides, FL can facilitate collaborative knowledge sharing and learning across organizations in the context of Industry 5.0. Organizations can collaborate to train machine learning models that capture and share domain-specific knowledge, insights, or best practices, while ensuring sensitive information is preserved. This approach can result in enhanced collaboration, innovation, and knowledge transfer among different stakeholders in Industry 5.0, promoting collective intelligence and driving advancements in the industrial sector. Moreover, FL can be employed to promote ethical and responsible use of AI in the context of Industry 5.0. Organizations can collaborate to train machine learning models that incorporate principles of fairness, transparency, and accountability, while actively avoiding bias, discrimination, or other unethical practices. This approach can lead to the development of more responsible and trustworthy AI systems in Industry 5.0, ensuring that AI technologies are used in a manner that aligns with ethical and societal standards. Table 2 summarizes how FL can contribute to both Industry 4.0 and 5.0 by linking the FL application scope with the agenda.

Table 2: Application of FL in enabling Industry 4.0 and 5.0

| Category | Application of FL | Agenda of Industry 4.0 | Agenda of Industry 5.0 |
|---|---|---|---|
| Focus | Creating more accurate and intelligent models by allowing multiple parties to | Focused on digitizing and optimizing the | Emphasis on the integration of human skills and expertise |

| | share their data and collaborate on the training process, while still maintaining data privacy and security. | manufacturing process through the integration of advanced technologies | with intelligent machines to create a more collaborative and flexible manufacturing environment. |
|---|---|---|---|
| Human-Machine Collaboration | Create a more seamless and secure collaboration between human workers and machines, by enabling real-time data sharing and analysis. | Integration of human workers with machines | More collaborative and flexible work environment through machines- human work integration |
| Flexibility | Creating a more flexible manufacturing process by allowing real-time updates to the training of machine learning models, and by enabling the use of distributed computing to scale up or down as needed. | Creating a more flexible manufacturing environment using advanced technologies. | To enhance this flexibility even further, by leveraging human skills and expertise to create a more agile and adaptive manufacturing process. |
| Customization | Creation of highly customized and personalized products by enabling the use of distributed data to train machine learning models that can be tailored to individual customer needs. | Creation of more customized products using 3D printing and other advanced manufacturing technologies. | Combining the strengths of both human workers and machines to create even more highly customized and personalized products. |
| Sustainability | By creating more energy-efficient and resource-efficient manufacturing processes, and by enabling distributed computing to reduce the carbon footprint of centralized data centers. | - | Emphasis on environmentally sustainable practices to create a sustainable and socially responsible manufacturing ecosystem |

## 5. Challenges in Adopting FL and Future Direction

This study reviewed 13 papers to identify the challenges of adopting FL in the manufacturing industry. The identified barriers were then categorized into four broad scopes of constraints, namely algorithm, organizational culture, security and privacy, and wireless devices. Table 3 presents these constraints along with future directions to address them. The table provides a comprehensive overview of the identified constraints and can serve as a useful guide for future research on the adoption of FL in the manufacturing industry.

Table 3: Barriers and potential solutions of adopting FL in the manufacturing industry

| Research | Type of Constraint | Barriers/Constraints | Future Directions |
|---|---|---|---|
| (Hu et al., 2018) | Algorithm | Need to be extended to multi-layer structures instead of a two-layer structure | The challenges of federated learning include the convergence of algorithms, especially for non-convex loss functions under limited communication and resources, and the trade-off between model performance and resource preservation. There is also a need for efficient approaches that reduce resource consumption, especially for low-powered devices such as IoT nodes. |
| (F. Liu et al., 2020) | Algorithm | More beneficial for the smaller dataset than the larger one in horizontal FL | |
| (Mowla et al., 2020) | Algorithm | Need to improve the reliability of the global updates in this architecture. | |
| (Saputra et al., 2019) | Algorithm | Need to be more stable and flexible. | |
| (Hegiste et al., 2022) | Algorithm | The accuracy achieved through FL is lower than deep-learning model | |
| (Müller et al., 2022) | Algorithm | A small number of available libraries | |
| (Rahman et al., 2021) | Algorithm | Addressing the research gap in granular variations in data characteristics, nonconvex objectives, and parallel Stochastic gradient descent. | |

| Reference | Category | Challenge | Solution |
|---|---|---|---|
| (Rahman et al., 2021) | Algorithm | A standard platform needs to be developed for a holistic ablation analysis of the different parts of an FL system. | |
| (Hu et al., 2018) | Cultural | Lack of quick model deployment to serve various industries | Successful implementation of federated learning in a centralized and hierarchical manufacturing organization requires a culture of trust, open communication, and willingness to share data. It also requires investment in infrastructure and technology for decentralized data processing and secure data transfer. |
| (Müller et al., 2022) | Cultural | Inevitable need for expert knowledge | |
| (Yang et al., 2019) | Cultural | Carrying out the business model for data alliance and the technical mechanism for federated learning should be done together. | |
| (Rahman et al., 2021) | Cultural | Trade-offs among accuracy, privacy, communication, and personalization level | |
| (Müller et al., 2022) | Security and Privacy | Data privacy and data security | To overcome these challenges, the secure aggregation algorithm has been proposed to aggregate encrypted local models, but participation disclosure is still an issue. Efficient algorithms that deliver comprehensive privacy analysis of deep learning, providing high performance and privacy protection without imposing additional computational burden, are needed. Additionally, mechanisms to prevent memorization, adversarial attacks, and membership inference attacks should be investigated and implemented to strengthen the security and privacy of Federated Learning. |
| (Fredrikson et al., 2015) | Security and Privacy | Model inversion attack by one of the participants | |
| (Nasr et al., 2019) | Security and Privacy | Membership inference attacks | |
| (Bagdasaryan et al., 2020) | Security and Privacy | An attacker being a part of a federated learning scheme may aim to influence the predictions of a trained model through poisoning or backdooring. | |
| (Ganju et al., 2018) | Security and Privacy | Property inference attacks aim to uncover the statistical characteristics of data sources. | |
| (Zheng et al., 2022) | Security and Privacy | The presence of malicious terminal devices during FL training can negatively impact accuracy and convergence by altering local learning model parameters. | |
| (Zheng et al., 2022) | Wireless Settings | The challenge of sparsity in FL due to limitations of wireless resources and noise in datasets can affect convergence and local model training. | A gradient-based sparsity scheme can be constructed by integrating communication resources, cleaning the dataset, and selecting appropriate computing power for training. |
| (Zheng et al., 2022) | Wireless Settings | The convergence performance of FL is significantly reduced due to the statistical heterogeneity present in many datasets. | A group of datasets can be formed by selecting terminal devices with a certain degree of trust. |
| (Zheng et al., 2022) | Wireless Settings | The allocation of resources during FL may impact cellular users and consume uplink communication resources. | A resource allocation mechanism based on game theory can effectively integrate resources and connects resource blocks with terminal devices. |
| (Rahman et al., 2021) | Wireless Settings | The challenge posed by systems heterogeneity in FL has been tackled by various algorithms, yet training may still be disrupted by wireless connectivity issues causing device dropouts. | New FL algorithms need to be designed to be more robust to handle a larger number of device dropouts. |

## 6. Conclusion
With the growing availability of time-series data, the manufacturing industry is increasingly moving towards real-time and sequential learning through edge devices and the cloud (Raihan & Ahmed, 2019). However, small-scale manufacturers still do not possess the capability of collecting enough data by themselves or analyzing them in the cloud for a real-time decision. Although FL offers a potential solution to address this issue without compromising data privacy, its implementation remains rare in the manufacturing industry. In this study, we attempted to provide a discussion of FL and its potential applications across the diversified manufacturing industry. We also discussed the applications FL from the context of Industry 4.0 and Industry 5.0. Moreover, the challenges associated with adopting FL in the manufacturing industry are extensively reviewed and categorized in this work. Based on the analysis, it can be concluded that most of the challenges relate to technical readiness, including software and hardware. Therefore, the focus for future research should be on addressing these technical challenges, while simultaneously promoting a shift in organizational culture to prepare the industry for the adoption of FL and drive improved outcomes.

## Acknowledgments
We would like to express our sincere gratitude to the Department of Industrial and Management Systems Engineering (IMSE) at West Virginia University (WVU) for their support in this research. Their resources, guidance, and expertise were invaluable in shaping the direction and scope of our study. We are also grateful for the encouragement and motivation provided by our professors and colleagues, which helped us to stay focused and committed throughout this project.

# Biographies

**Farzana Islam** is a master's student in the Department of Industrial and Management Systems Engineering at West Virginia University. She graduated from Bangladesh University of Engineering and Technology with a major in Industrial Engineering and has two years of work experience in the manufacturing and service industries. She previously worked in the service industry as a project coordinator for human resources transformation projects such as ERP implementation and competency frameworks roll-out projects. She has led projects that promote sustainability through the reduction of carbon footprint generated by the supply chain network in her role as a planning officer in the manufacturing industry. She has also received several best paper awards in the tracks of sustainability, lean six-sigma, optimization, and supply chain at the Industrial Engineering and Operation Management conferences. Her current research focuses on adding new dimensions to the sustainable approaches used by industries by leveraging her prior experiences and education.

**Ahmed Shoyeb Raihan** has completed his undergraduate studies in Industrial and Production Engineering (IPE) from Bangladesh University of Engineering and Technology (BUET) in 2019. In the same year, he joined BGMEA University of Fashion and Technology (BUFT) as a lecturer in the Department of Industrial Engineering (IE). In 2017, he achieved the 'Certified Supply Chain Analyst (CSCA)' certification from ISCEA. Currently, he is doing his Ph.D. in Industrial Engineering from West Virginia University. He is also working as a Graduate Research Assistant in the System Analytics Lab at West Virginia University. His field of research encompasses machine learning, data science, supply chain and logistics, sustainability, operations research, and optimization.

**Imtiaz Ahmed** received his B.Sc. and M.Sc. degrees in Industrial & Production engineering from the Bangladesh University of Engineering and Technology, Bangladesh, in 2012 and 2014, respectively. He served as a faculty member with the Bangladesh University of Engineering & Technology for three years and was actively involved in teaching, research and consultancy works. He also served as an affiliated faculty member in different universities in Bangladesh. Imtiaz received his PhD. in Industrial Engineering from Texas A&M University in 2020. Before joining WVU, he worked as a postdoctoral researcher in the Industrial and Systems Engineering Department, Texas A&M University for one year. He is an active member of INFORMS and IISE and actively participates in these annual meetings. He won several competitive awards in these conferences. His research interests are in data science, machine learning, quality control and inventory management.